\documentclass[runningheads]{llncs}
\usepackage[T1]{fontenc}
\usepackage{graphicx,verbatim}
\usepackage{amsmath,amssymb}
\usepackage{booktabs}
\usepackage{multirow}

\usepackage{xcolor}

\definecolor{revisionblue}{RGB}{0,70,180}

\newcommand{\rev}[1]{{\color{black}#1}}

\newenvironment{revblock}
  {\begingroup\color{black}}
  {\endgroup}

\begin{document}

\title{
Deterministic Hallucination Detection in Medical VQA via Confidence-Evidence Bayesian Gain}
\titlerunning{CEBaG for Medical VQA Hallucination Detection}

\author{Mohammad Asadi\inst{1} \and
Tahoura Nedaee\inst{2} \and
Jack W. O'Sullivan\inst{3,4} \and
Euan Ashley\inst{3,4} \and
Ehsan Adeli\inst{4,5,6}}
\authorrunning{M. Asadi et al.}
\institute{Department of Electrical Engineering, Stanford University, CA, USA\\
\email{masadi@stanford.edu}\and
Department of Biology, Stanford University, CA, USA \and
Division of Cardiology, Department of Medicine, Stanford University, CA, USA \and
Department of Biomedical Data Science, Stanford University, CA, USA \and
Department of Computer Science, Stanford University, CA, USA \and
Department of Psychiatry and Behavioral Sciences, Stanford University, CA, USA}


\maketitle

\begin{abstract}
Multimodal large language models (MLLMs) have shown strong potential for medical Visual Question Answering (VQA), yet they remain prone to hallucinations, defined as generating responses that contradict the input image, posing serious risks in clinical settings. Current hallucination detection methods, such as Semantic Entropy (SE) and Vision-Amplified Semantic Entropy (VASE), require 10 to 20 stochastic generations per sample together with an external natural language inference model for semantic clustering, making them computationally expensive and difficult to deploy in practice.
We observe that
hallucinated responses exhibit a distinctive signature directly in the
model's own log-probabilities: inconsistent token-level confidence and
weak sensitivity to visual evidence. Based on this observation, we propose
\textbf{C}onfidence-\textbf{E}vidence \textbf{Ba}yesian \textbf{G}ain (CEBaG), a deterministic
\rev{answer-level} hallucination detection method \rev{for short-form medical
VQA} that requires no stochastic sampling, no external models, and no
task-specific hyperparameters. CEBaG combines
two complementary signals: \emph{token-level predictive variance}, which
captures inconsistent confidence across response tokens, and \emph{evidence magnitude}, which measures how much the image shifts per-token
predictions relative to text-only inference. Evaluated across four medical MLLMs and three VQA benchmarks (16 experimental settings), CEBaG achieves the highest AUC in 13 of 16 settings and improves over VASE by 8~AUC points on average, while being fully deterministic and self-contained. https://github.com/masadi-99/CEBaG.

\end{abstract}

\section{Introduction}

Medical Visual Question Answering (VQA), powered by multimodal large language models (MLLMs), enables the interpretation of medical images in response to natural language queries, supporting tasks from abnormality detection to differential diagnosis~\cite{xiao2024survey,liu2024gemex}. However, MLLMs are susceptible to \emph{hallucinations}, defined as responses that misinterpret or contradict the input image, posing substantial risks including misdiagnoses, inappropriate treatments, and diminished clinician trust~\cite{huang2023survey,liu2024hallsurvey}. Despite efforts in data optimization~\cite{yu2024hallucidoctor}, training~\cite{jiang2024hacl}, and decoding refinements~\cite{wang2024icd}, hallucinations remain a persistent challenge, highlighting the need for effective detection methods.

Recent approaches to hallucination detection include detector fine-tuning~\cite{gunjal2024detecting}, cross-checking~\cite{yu2024hallucidoctor,cohen2023lmvslm}, visual evidence verification~\cite{yin2024woodpecker}, and uncertainty estimation~\cite{li2024referencefree,chen2024inside,farquhar2024se,zhang2024vluncertainty}. Among these, uncertainty estimation stands out for its simplicity, as it uses only the model itself without additional annotated data or external knowledge bases~\cite{huang2023survey,liu2024hallsurvey}. Semantic Entropy (SE)~\cite{farquhar2024se} quantifies uncertainty by computing entropy over semantically clustered responses. Vision-Amplified Semantic Entropy (VASE)~\cite{liao2025vase} extends SE by contrasting semantic distributions from original and augmented images, amplifying the influence of visual input. Other contrastive methods such as VCD~\cite{leng2024vcd} and LCD~\cite{manevich2024lcd} leverage output differences under perturbed inputs, while Semantic Entropy Probes~\cite{kossen2024seprobes} train a probe on internal activations, requiring task-specific data unlike our training-free approach. A common thread in these methods is their reliance on stochastic sampling, external models, or additional training, which raises the question of whether the uncertainty signal they seek is accessible through simpler means.

\begin{figure}[!ht]
\centering
\includegraphics[width=\columnwidth]{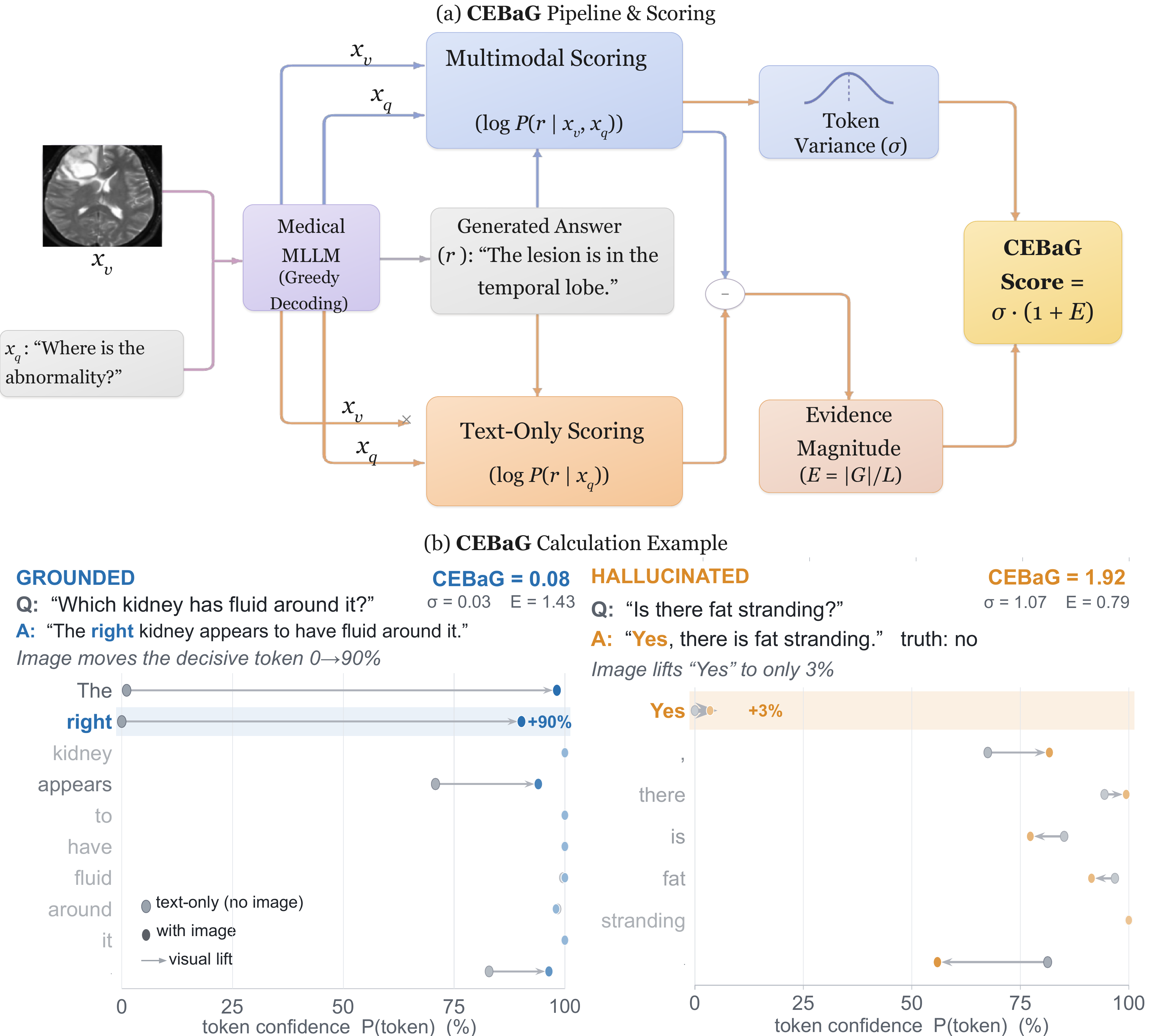}
\caption{Overview of CEBaG for hallucination detection in medical VQA. (a)~The CEBaG pipeline: the model generates answer~$r$, scored with and without the image to yield two signals: token-level predictive variance~$\sigma$ and evidence magnitude $E{=}|G|/L$. CEBaG combines these as $\sigma\cdot(1{+}E)$. (b)~Per-token probabilities ($100\,P(\text{token})$) with vs.\ without the image, and CEBaG scores for a grounded and a hallucinated example.}
\label{fig:pipeline}
\end{figure}

These methods also face a concrete efficiency bottleneck. SE requires $M$ stochastic generations (typically $M{=}10$) with high-temperature sampling, followed by pairwise semantic equivalence checks via an external NLI model~\cite{he2020deberta}. VASE doubles this to $2M$ generations plus $O(M^2)$ NLI comparisons per input, making it prohibitive for clinical deployment where real-time feedback is essential. We propose \textbf{Confidence-Evidence Bayesian Gain (CEBaG)}, which replaces stochastic sampling with deterministic log-probability analysis. CEBaG combines two complementary signals: \emph{token-level predictive variance} (the standard deviation of per-token log-probabilities, elevated when the model fabricates content) and \emph{evidence magnitude} (the absolute per-token log-probability shift when the image is present versus absent, capturing the strength of visual grounding). The resulting score, $\sigma \cdot (1 + |G|/L)$, is parameter-free, requires only one generation and two scoring passes, uses no external models, and is fully deterministic (Fig.~\ref{fig:pipeline}). \begin{revblock}
We position CEBaG specifically for answer-level hallucination detection in short-form
medical VQA. Its two signals, the sequence-level evidence gain $G$ and the
global standard deviation of token log-probabilities, operate over the entire
response and thus capture whether an answer, taken as a whole, is grounded in
the image. This makes CEBaG well-suited to flagging holistic answer--image
consistency in concise medical VQA.
\end{revblock}

Our contributions are: (1)~We propose CEBaG, a deterministic \rev{answer-level} hallucination detection method requiring three forward passes and no external models, compared to 20+ stochastic generations and an NLI model for SE/VASE. (2)~We conduct an extensive evaluation on four MLLMs (MedGemma-4b, MedGemma-1.5-4b, HuatuoGPT-Vision-7B, LLaVA-Med-v1.5-7B) and three datasets (VQA-RAD, SLAKE, PathVQA), 16 settings in total.
(3)~CEBaG consistently outperforms SE, VASE, and RadFlag~\cite{sambara2024radflag}, achieving the best AUC in 13 of 16 settings and improving over VASE by 8~AUC points on average.

\section{Method}


Given a medical image $x_v$, a textual question $x_q$, and a medical MLLM $f$, the model generates a response $r = f(x_v, x_q)$ via greedy or low-temperature decoding. Our goal is to determine whether $r$ is a hallucination (a response that contradicts the visual evidence in $x_v$) without access to ground-truth answers. For evaluation, we follow prior work~\cite{liao2025vase,farquhar2024se} and use the GREEN model~\cite{ostmeier2024green} to produce reference-based ground-truth labels.

\subsection{Evidence Gain as Pointwise Mutual Information}

The key quantity in CEBaG is the \emph{evidence gain}: how much the model's belief in its own answer changes when it observes the image. Formally, we interpret the text-only probability $P(r \mid x_q)$ as the model's \textbf{language prior}, i.e., its belief about the answer based solely on the question and its pre-trained knowledge. The multimodal probability $P(r \mid x_v, x_q)$ represents the \textbf{posterior} belief after updating with the visual evidence $x_v$. We use ``prior'' and ``posterior'' in the information-theoretic sense: $P(r \mid x_q)$ represents the model's belief before observing visual evidence, and $P(r \mid x_v, x_q)$ its updated belief after conditioning on the image. This is distinct from Bayesian inference over model parameters; rather, $G$ captures how the image updates the model's predictive distribution for a fixed set of parameters.
The sequence log-probabilities are defined as:
\begin{equation}
\log P(r \mid x_v, x_q) = \sum\nolimits_{j=1}^{L} \log P(r_j \mid r_{<j},\, x_v,\, x_q),
\label{eq:logp_img}
\end{equation}
\begin{equation}
\log P(r \mid x_q) = \sum\nolimits_{j=1}^{L} \log P(r_j \mid r_{<j},\, x_q).
\label{eq:logp_text}
\end{equation}

The \textbf{evidence gain} $G$ is the difference between the posterior and the prior:
\begin{equation}
G = \log P(r \mid x_v, x_q) - \log P(r \mid x_q).
\label{eq:gain}
\end{equation}

By Bayes' rule, $P(r \mid x_v, x_q) = \frac{P(x_v \mid r, x_q) P(r \mid x_q)}{P(x_v \mid x_q)}$. Substituting this into Eq.~\eqref{eq:gain}: 
{\small\begin{equation}
{G = \log \left( \frac{P(x_v \mid r, x_q) P(r \mid x_q)}{P(x_v \mid x_q)} \right) - \log P(r \mid x_q) = \log P(x_v \mid r, x_q) - \log P(x_v \mid x_q).}
\label{eq:pmi}
\end{equation}}
Since $\log P(x_v \mid x_q)$ is constant with respect to the response $r$, $G$ is directly proportional to $\log P(x_v \mid r, x_q)$, the log-likelihood of the image given the response. Thus, $G$ is the \textbf{Pointwise Mutual Information (PMI)} between the response and the image (conditioned on the question).
A large positive $G$ implies that the response makes the observed image highly probable (strong grounding), while a negative $G$ suggests the response contradicts the visual evidence.

The \textbf{evidence magnitude} is a length-normalized visual influence measure:
\begin{equation}
E = \frac{|G|}{L},
\label{eq:evidence_mag}
\end{equation}
which captures the average absolute shift in belief per token. We use the magnitude $|G|$ because a strong visual influence (increasing or decreasing token probabilities) indicates attention to the image, whereas a lack of influence ($G \approx 0$) suggests the model is relying entirely on its language prior.

\subsection{Confidence-Evidence Bayesian Gain}

We propose CEBaG, a scoring function that combines the model's internal uncertainty with its sensitivity to visual evidence. We define the \textbf{token-level predictive variance}:
$\sigma = \mathrm{std}_{j=1}^{L}\!\bigl[\log P(r_j \mid r_{<j},\, x_v,\, x_q)\bigr],$
which measures the consistency of the model's confidence across the generated sequence. High $\sigma$ indicates ``confidence spikes'' on function words interspersed with low-confidence content tokens, a pattern characteristic of hallucination.
The CEBaG score is defined as:
\begin{equation}
\text{CEBaG} = \sigma \cdot \bigl(1 + E\bigr).
\label{eq:CEBaG}
\end{equation}
This formulation can be interpreted as \textbf{posterior uncertainty weighted by visual information gain}.
Hallucinations typically arise in two scenarios: (1) The model is uncertain about the content ($\sigma$ is high); or (2) The visual evidence causes a large shift in belief ($E$ is high) but fails to resolve the uncertainty (or contradicts the language prior).

The multiplicative term $(1+E)$ amplifies the base uncertainty $\sigma$ when the response is highly sensitive to the image, flagging cases where the model is ``struggling'' to reconcile visual evidence with its language prior.
\begin{revblock}
This amplification is realized as a multiplicative \emph{gain} on $\sigma$ rather than an independent additive term. The gain is monotone in both signals and preserves $\sigma$ as its base, such that the score reduces to $\sigma$ when the image leaves the prediction unchanged ($E{=}0$), so visual evidence can only sharpen, never override, the uncertainty estimate. Because it multiplies rather than adds, it also requires no coefficient to reconcile the differing scales of $\sigma$ and $E$.
\end{revblock}

Importantly, this formulation is \textbf{hyperparameter-free}, requiring no hyperparameter tuning or normalization, making it robust across different models and datasets. Unlike SE and VASE, which require configuring the number of generations $M$, sampling temperature $T$, and (for VASE) the amplification ratio $\alpha$, CEBaG applies the same fixed formula across all models and datasets. 


\section{Experiments}

\noindent\textbf{Datasets:} We evaluate on three medical VQA benchmarks: \textit{VQA-RAD}~\cite{lau2018vqarad} (451 test samples, including 200 open-ended), a radiology VQA dataset covering chest X-rays, CT, and MRI; \textit{SLAKE}~\cite{liu2021slake}, a bilingual medical VQA dataset with semantically labeled knowledge (1061 test samples); and \textit{PathVQA}~\cite{he2020pathvqa} (1000 test samples), covering pathology images. For VQA-RAD, we report performance on both open-ended questions and all questions (including yes/no), following~\cite{liao2025vase}.

\noindent\textbf{Models:} Four medical MLLMs spanning two architecture families: \textit{MedGemma-4b-it} and \textit{MedGemma-1.5-4b-it}~\cite{sellergren2025medgemma} (encoder--decoder), and \textit{HuatuoGPT-Vision-7B}~\cite{chen2024huatuogpt} and \textit{LLaVA-Med-v1.5-Mistral-7B}~\cite{li2024llavamed} (LLaVA-style decoder-only). 

\noindent\textbf{Ground truth and metrics:} Following the evaluation protocol established by VASE~\cite{liao2025vase}, we use the GREEN model~\cite{ostmeier2024green} to generate reference-based ground-truth labels: a response is labeled as hallucinated if its GREEN score falls below 1.0. We adopt this protocol to ensure direct comparability with prior work and use the same metrics (AUC and AUG) as~\cite{liao2025vase}.

\noindent\textbf{Baselines.} We compare CEBaG against four state-of-the-art hallucination detection methods: (1) \textit{AvgProb}~\cite{li2024referencefree}, which uses the average token probability in the sequence; (2) \textit{Semantic Entropy (SE)}~\cite{farquhar2024se}, which measures uncertainty over semantic clusters of generated answers; (3) \textit{Vision-Amplified Semantic Entropy (VASE)}~\cite{liao2025vase}, which enhances SE by generating additional answers conditioned on visually augmented images; and (4) \textit{RadFlag}~\cite{sambara2024radflag}, which checks for consistency between the generated answer and the findings in the radiology report. We adhere to the official implementations and hyperparameters for all baselines. For SE and VASE, we use 10 generations (plus 10 augmented for VASE) and a temperature of 1.0, as specified in their respective papers.

\noindent\textbf{Implementation details.} For CEBaG, the generated answer $r$ is produced via greedy decoding (temperature $T{=}0.1$). The two scoring passes (Eqs.~\ref{eq:logp_img}--\ref{eq:logp_text}) use teacher-forced forward passes; $\sigma$ is computed from the same forward pass as $\log P(r \mid x_v, x_q)$ at no additional cost. For the text-only pass, visual input is removed at the interface level: for MedGemma, no image is included in the chat message and no pixel values are passed to the model; for LLaVA-style models (LLaVA-Med, HuatuoGPT), the \texttt{<image>} token is removed from the prompt template and the processor is called without image input. This ensures a clean text-only baseline across all architectures. All experiments were conducted on 8 NVIDIA H100 GPUs. Following the evaluation protocol of SE~\cite{farquhar2024se} and VASE~\cite{liao2025vase}, we report point estimates of AUC and AUG. Being deterministic, CEBaG avoids the baselines' sampling variance, giving exact per-answer scores.

\subsection{Main Results}

\begin{table}[t]
\caption{Hallucination detection performance: AUC~(\%)$\uparrow$ and AUG~(\%)$\uparrow$ across four medical MLLMs and three VQA datasets. Best results per row are in \textbf{bold}; second best are \underline{underlined}. CEBaG achieves the best AUC in 13 of 16 settings.}\label{tab:main}
\centering
\fontsize{8}{9.5}\selectfont
\setlength{\tabcolsep}{2pt}
\begin{tabular}{l cc cc cc cc cc}
\toprule
& \multicolumn{2}{c}{AvgProb~\cite{li2024referencefree}} & \multicolumn{2}{c}{SE~\cite{farquhar2024se}} & \multicolumn{2}{c}{VASE~\cite{liao2025vase}} & \multicolumn{2}{c}{RadFlag~\cite{sambara2024radflag}} & \multicolumn{2}{c}{\textbf{CEBaG (Ours)}} \\
\cmidrule(lr){2-3} \cmidrule(lr){4-5} \cmidrule(lr){6-7} \cmidrule(lr){8-9} \cmidrule(lr){10-11}
Dataset & AUC & AUG & AUC & AUG & AUC & AUG & AUC & AUG & AUC & AUG \\
\midrule
\multicolumn{11}{l}{\textit{MedGemma-4b-it}~\cite{sellergren2025medgemma}} \\
\quad VQA-RAD (Open) & 39.9 & 48.0 & \textbf{61.9} & \textbf{65.3} & 60.8 & 63.3 & 58.4 & 63.6 & \underline{61.0} & \underline{64.8} \\
\quad VQA-RAD (All) & 42.9 & 54.6 & 52.2 & 58.0 & \underline{57.6} & \underline{59.9} & 54.7 & 58.9 & \textbf{60.8} & \textbf{64.6} \\
\quad SLAKE & 35.3 & 52.9 & 54.9 & 64.2 & \underline{62.6} & \underline{70.1} & 57.0 & 66.2 & \textbf{71.9} & \textbf{75.9} \\
\quad PathVQA & 46.4 & 45.1 & 58.0 & 48.9 & \underline{61.5} & \underline{49.6} & 56.8 & 47.4 & \textbf{71.9} & \textbf{59.0} \\
\midrule
\multicolumn{11}{l}{\textit{MedGemma-1.5-4b-it}~\cite{sellergren2025medgemma}} \\
\quad VQA-RAD (Open) & 69.1 & 47.4 & 53.2 & 37.2 & \underline{60.1} & \underline{40.3} & 54.5 & 39.3 & \textbf{72.9} & \textbf{47.2} \\
\quad VQA-RAD (All) & 63.6 & 56.2 & 53.4 & 48.6 & \underline{54.8} & \underline{51.5} & 54.3 & 46.3 & \textbf{66.9} & \textbf{58.2} \\
\quad SLAKE & 57.4 & 52.0 & \underline{57.7} & 51.9 & \textbf{59.5} & \underline{52.9} & 57.2 & \textbf{53.0} & 57.5 & 52.0 \\
\quad PathVQA & 65.9 & 48.9 & 56.8 & 40.9 & \underline{60.1} & \underline{45.7} & 56.3 & 41.6 & \textbf{66.6} & \textbf{48.8} \\
\midrule
\multicolumn{11}{l}{\textit{HuatuoGPT-Vision-7B}~\cite{chen2024huatuogpt}} \\
\quad VQA-RAD (Open) & 43.7 & 48.1 & \underline{64.1} & \underline{62.8} & 61.7 & 59.6 & 59.9 & 58.4 & \textbf{70.4} & \textbf{63.1} \\
\quad VQA-RAD (All) & 39.3 & 52.2 & 58.3 & 65.1 & 58.9 & \underline{66.1} & \underline{60.0} & 63.5 & \textbf{67.7} & \textbf{67.6} \\
\quad SLAKE & 30.5 & 38.4 & 57.7 & 54.9 & 59.9 & \underline{58.0} & \underline{60.6} & 54.2 & \textbf{67.0} & \textbf{64.2} \\
\quad PathVQA & 30.5 & 28.7 & 53.6 & 38.2 & 53.2 & \underline{39.5} & \underline{55.5} & 38.9 & \textbf{82.0} & \textbf{55.6} \\
\midrule
\multicolumn{11}{l}{\textit{LLaVA-Med-v1.5-Mistral-7B}~\cite{li2024llavamed}} \\
\quad VQA-RAD (Open) & 62.9 & 38.6 & 63.8 & \textbf{40.4} & \textbf{68.4} & \underline{40.2} & \underline{65.8} & 40.1 & 65.8 & 39.1 \\
\quad VQA-RAD (All) & 58.8 & 48.7 & 58.1 & \textbf{50.7} & 57.6 & 49.3 & 56.8 & 48.1 & \textbf{62.1} & \underline{50.0} \\
\quad SLAKE & 59.9 & 50.0 & 57.9 & 50.2 & 57.4 & \underline{50.8} & \underline{58.7} & 50.7 & \textbf{63.8} & \textbf{52.3} \\
\quad PathVQA & 75.1 & 52.0 & 57.4 & 41.3 & 60.4 & 41.7 & 58.8 & 39.7 & \textbf{77.6} & \textbf{53.1} \\
\midrule
\textbf{Average} & 51.3 & 47.6 & 57.4 & 51.2 & 59.6 & 52.4 & 57.8 & 50.6 & \textbf{67.9} & \textbf{57.2} \\
\bottomrule
\end{tabular}
\end{table}

Table~\ref{tab:main} presents the hallucination detection performance across all 16 experimental settings. CEBaG achieves the best AUC in \textbf{13 of 16} settings and ranks in the top two in 15 of~16. On average, CEBaG attains 67.9\% AUC, outperforming VASE (59.6\%), SE (57.4\%), and RadFlag (57.8\%) by substantial margins, with an improvement of +8.2~AUC points over the previous state-of-the-art VASE. Significantly, CEBaG achieves this without any hyperparameter tuning.

The gains are particularly pronounced on larger datasets: on PathVQA, CEBaG outperforms VASE by +10.5 (MedGemma), +6.5 (MedGemma-1.5), +28.8 (HuatuoGPT), and +17.2 (LLaVA-Med) AUC points. On SLAKE, the improvements are similarly large, with CEBaG surpassing VASE by +9.3 (MedGemma) and +7.1 (HuatuoGPT). On the smaller VQA-RAD datasets, the improvements remain consistent across most models, with CEBaG achieving the best AUC in 6 of 8 VQA-RAD settings. Furthermore, our method maintained stable detection performance across GREEN thresholds between 0.4 and 0.8, confirming its insensitivity to specific ground-truth boundaries.

\subsection{Efficiency Analysis}

\begin{table}[t]
\caption{Computational cost and performance. Despite an order of magnitude fewer forward passes, CEBaG achieves the highest average AUC and AUG.}\label{tab:efficiency}
\centering
\fontsize{8}{9.5}\selectfont
\begin{tabular}{lcccccc}
\toprule
Method & Forward passes & External model & Deterministic & Hyperparams & $\overline{AUC}$ & $\overline{AUG}$ \\
\midrule
AvgProb~\cite{li2024referencefree} & 1 gen. & None & Yes & None & 51.3 & 47.6 \\
SE~\cite{farquhar2024se} & $M$ gen.\ + NLI & DeBERTa-MNLI & No & $M$, $T$ & 57.4 & 51.2 \\
VASE~\cite{liao2025vase} & $2M$ gen.\ + NLI & DeBERTa-MNLI & No & $M$, $T$, $\alpha$ & 59.6 & 52.4 \\
RadFlag~\cite{sambara2024radflag} & $M$ gen.\ + NLI & DeBERTa-MNLI & No & $M$, $T$ & 57.8 & 50.6 \\
\textbf{CEBaG (Ours)} & 1 gen.\ + 2 evals & None & Yes & None & \textbf{67.9} & \textbf{57.2} \\
\bottomrule
\end{tabular}
\end{table}

Table~\ref{tab:efficiency} compares the computational requirements and resulting average performance for each method. With the standard setting $M{=}10$, SE requires 10 autoregressive generations plus $O(M^2){=}O(100)$ pairwise entailment comparisons per sample. VASE doubles the generation cost to 20 passes plus the same entailment overhead. RadFlag also requires $M$ generations and entailment checks. All three methods additionally require loading and running a DeBERTa-v2-xlarge-MNLI model alongside the target MLLM, and all involve hyperparameters ($M$, sampling temperature~$T$, and for VASE, the vision-amplified ratio~$\alpha$).

In contrast, CEBaG requires exactly \textbf{three forward passes}: one autoregressive generation and two teacher-forced scoring passes (with and without the image). No external model is needed, and no hyperparameters require tuning. The scoring passes are substantially cheaper than generation as they process fixed token sequences without autoregressive decoding. Despite its simplicity and efficiency, CEBaG achieves the highest average AUC (67.9\%) and AUG (57.2\%) across all settings, outperforming the computationally expensive baselines.

\subsection{Ablation Study}

We compare four variants: $\sigma$~only (token-level variance); $E$~only (evidence magnitude $|G|/L$); CEBaG, the parameter-free $\sigma\cdot(1+E)$; and CEBaG$_\lambda$, \rev{an oracle that, per model-dataset pair, min-max normalizes $G$ and a penalty term and selects the gain sign, penalty term, and weight $\lambda\in[-5,5]$ maximizing AUC, an in-sample configuration unattainable without test labels.}

\begin{table}[t]
\caption{Ablation study: AUC~(\%)$\uparrow$ of CEBaG components. $\sigma$~only uses token-level variance alone; $E$~only uses absolute per-token gain alone; CEBaG combines both (parameter-free); CEBaG$_\lambda$ adds a tunable weight (upper bound).}\label{tab:ablation}
\centering
\fontsize{8}{9.5}\selectfont
\setlength{\tabcolsep}{3.5pt}
\begin{tabular}{l cccc cccc}
\toprule
& \multicolumn{4}{c}{AUC (\%)} & \multicolumn{4}{c}{AUG (\%)} \\
\cmidrule(lr){2-5} \cmidrule(lr){6-9}
Setting & $\sigma$ & $E$ & CEBaG & CEBaG$_\lambda$ & $\sigma$ & $E$ & CEBaG & CEBaG$_\lambda$ \\
\midrule
\multicolumn{9}{l}{\textit{MedGemma-4b-it}} \\
\quad SLAKE & 71.1 & 51.6 & \underline{71.9} & \textbf{73.5} & \underline{74.8} & 64.9 & \textbf{75.9} & 74.6 \\
\quad PathVQA & 67.8 & 63.4 & \underline{71.9} & \textbf{77.5} & \underline{55.6} & 57.0 & \textbf{59.0} & 61.7 \\
\midrule
\multicolumn{9}{l}{\textit{MedGemma-1.5-4b-it}} \\
\quad VQA-RAD (Open) & 68.2 & 63.9 & \underline{72.9} & \textbf{75.7} & 44.7 & 40.8 & \underline{47.2} & \textbf{47.0} \\
\quad PathVQA & 57.9 & \underline{70.8} & 66.6 & \textbf{77.8} & 42.6 & \underline{52.2} & 48.8 & \textbf{56.3} \\
\midrule
\multicolumn{9}{l}{\textit{HuatuoGPT-Vision-7B}} \\
\quad PathVQA & \underline{80.6} & 67.3 & \textbf{82.0} & 82.5 & \underline{55.3} & 48.3 & \textbf{55.6} & 55.5 \\
\quad VQA-RAD (All) & \underline{68.0} & 53.8 & 67.7 & \textbf{68.6} & \textbf{67.5} & 59.6 & \underline{67.6} & 66.9 \\
\midrule
\multicolumn{9}{l}{\textit{LLaVA-Med-v1.5-Mistral-7B}} \\
\quad VQA-RAD (All) & \underline{62.9} & 52.0 & 62.1 & \textbf{66.2} & \underline{50.8} & 45.6 & 50.0 & \textbf{54.0} \\
\quad SLAKE & \underline{65.2} & 51.8 & 63.8 & \textbf{68.8} & \underline{52.7} & 46.8 & 52.3 & \textbf{55.3} \\
\midrule
\textbf{Average (all 16)} & 67.0 & 59.0 & \underline{67.9} & \textbf{71.9} & 53.8 & 50.5 & \underline{57.2} & \textbf{58.8} \\
\bottomrule
\end{tabular}
\end{table}

Table~\ref{tab:ablation} reports the ablation; the bottom row averages all 16 settings. 1- Token-level variance dominates: $\sigma$-only achieves 67.0\% average AUC, above all baselines (SE~57.4\%, VASE~59.6\%, RadFlag~57.8\%). 2- Evidence magnitude is complementary: while $E$~only (59.0\%) trails $\sigma$~only, CEBaG (67.9\%) outperforms it in \rev{more than half} of \rev{the} settings, with \rev{notable} gains on MedGemma-1.5 VQA-RAD~Open (+4.7) and MedGemma PathVQA (+4.1). \rev{In the rest, $(1+E)$ lowers AUC, usually marginally ($-0.3$ on HuatuoGPT VQA-RAD~All, $-1.4$ on LLaVA-Med SLAKE). Because $\sigma$ and $E$ combine without normalization, their relative influence varies across models; CEBaG keeps this factor because it raises the average and needs no tuning.} 3- \rev{The tuned upper bound does not transfer. CEBaG$_\lambda$ reaches 71.9\% average AUC, 4.0 points above CEBaG, entirely from per-setting $\lambda$ selection: the optimal $\lambda$ is unstable, ranging from $-5$ to $+2.75$. The best single $\lambda$ recovers under one of these four points, and the tuned value is unattainable without test-label access.}

\subsection{Robustness and Signal Analysis}

\begin{figure}[t]
\centering
\includegraphics[width=0.7\textwidth]{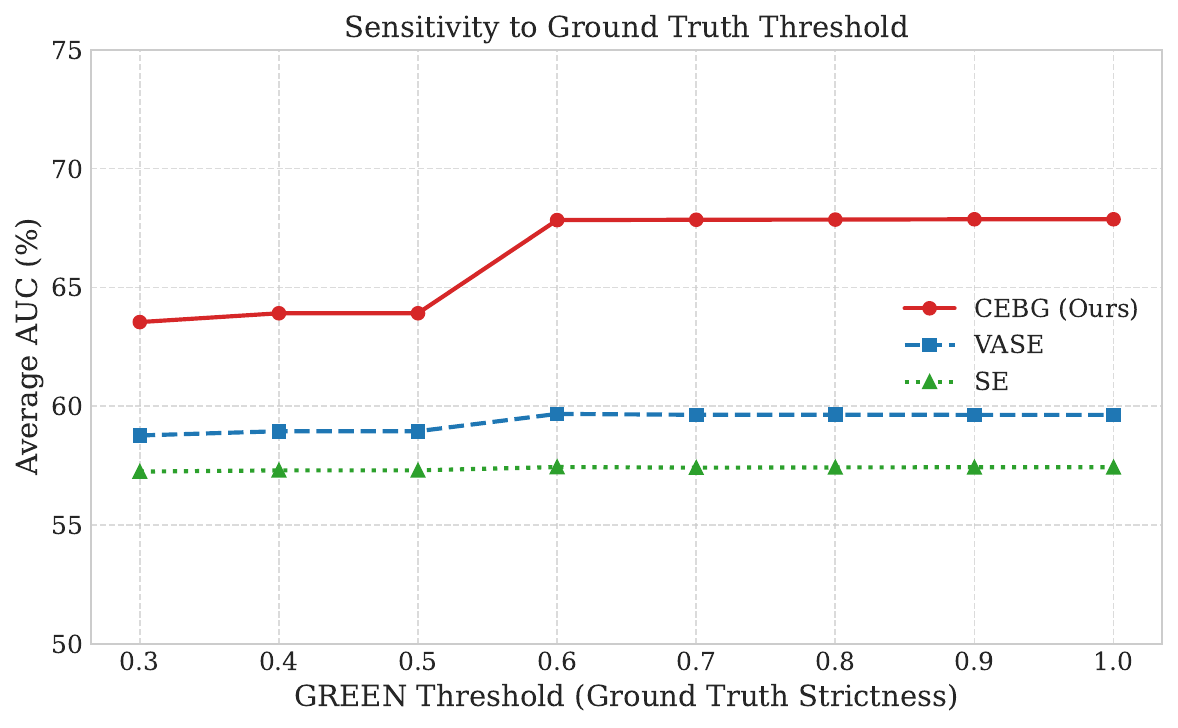}
\caption{Sensitivity to ground-truth threshold. CEBaG maintains high AUC across all GREEN thresholds (0.3--1.0), whereas VASE and SE degrade significantly at lower thresholds. This indicates CEBaG is robust to the strictness of the hallucination definition.}
\label{fig:sensitivity}
\end{figure}

\paragraph{Robustness to ground-truth definition.}
The definition of hallucination can be subjective. We evaluate the sensitivity of each method to the strictness of the ground-truth label by varying the GREEN score threshold from 0.3 to 1.0. Figure~\ref{fig:sensitivity} shows the average AUC across all datasets. CEBaG maintains a consistent AUC ($\sim$70--80\%) across all thresholds, demonstrating that its signal (internal uncertainty weighted by visual influence) aligns robustly with various levels of judgments.

\begin{table}[t]
\caption{Correlation of model log-probabilities: Real image vs.\ Noisy image (VASE protocol) and Real image vs.\ Text-only (CEBaG protocol). High correlation between real and noisy images suggests that noise injection (as used in VASE) may not sufficiently disrupt the model's visual prior, limiting the contrastive signal. Text-only priors provide a more distinct reference point.}\label{tab:correlation}
\centering
\fontsize{8}{9.5}\selectfont
\begin{tabular}{lcc}
\toprule
Model & Corr(Real, Noisy) & Corr(Real, Text-only) \\
\midrule
HuatuoGPT-Vision-7B & 0.995 & 0.974 \\
LLaVA-Med-v1.5-Mistral-7B & 0.999 & 0.957 \\
MedGemma-1.5-4b-it & 0.998 & 0.975 \\
MedGemma-4b-it & 0.938 & 0.777 \\
\bottomrule
\end{tabular}
\end{table}

\paragraph{Why text priors outperform noisy image priors.}
Methods like VASE~\cite{liao2025vase} and VCD~\cite{leng2024vcd} rely on contrasting the model's output given a real image versus a distorted (noisy) image. However, Table~\ref{tab:correlation} reveals that for modern medical MLLMs, the log-probabilities under noisy images (using VASE's noise protocol: rotation, crop, blur) are extremely highly correlated with those under real images ($r > 0.99$ for three models). This suggests that the model's visual representation is robust to these perturbations, making the noisy image a poor reference for ``uninformed'' belief. In contrast, CEBaG's text-only prior (removing visual tokens entirely) yields lower correlations ($0.95$--$0.97$), providing a more distinct baseline against which to measure visual influence. This explains why CEBaG's evidence gain $G$ (PMI) offers a stronger detection signal than VASE's noise-based contrast.

\section{Conclusion}

We introduced CEBaG, a deterministic \rev{answer-level} hallucination detection method for medical VQA combining token-level predictive variance and evidence magnitude. Unlike sampling-based approaches (SE, VASE) that need 10--20 stochastic generations and an external NLI model, CEBaG uses only three forward passes. Across four medical MLLMs and 16 settings, it achieves the best AUC in 13 and surpasses state-of-the-art VASE by 8~AUC points on average at much lower cost. Its efficiency, simplicity, and zero-configuration design make it practical for real-time clinical deployment; its main limitation is requiring white-box access to log-probabilities.
Future work can extend CEBaG to black-box settings via output perturbation and to multi-turn clinical dialogue. \begin{revblock}Because both signals aggregate over the full response, CEBaG operates at the whole-answer level; extending the per-token gain into a span-level signal for long-form outputs like report generation is a natural next direction. GREEN-generated labels, used for comparability with prior work, also make validation on a clinician-annotated subset an important next step.\end{revblock}

\noindent\textbf{Acknowledgment: }
This work was partially supported by the NIH Grants AG089169, AG084471, Stanford HAI Hoffman-Yee Award \& GCP Credits, and UIT. M. Asadi is supported by the Amazon AI PhD and the Stanford HAI Graduate Fellowships.

\noindent\textbf{\discintname}
The authors have no competing interests to declare.

\end{document}